\documentclass[review,authoryear,3p,10pt]{elsarticle}

\makeatletter
\def\ps@pprintTitle{%
 \let\@oddhead\@empty
 \let\@evenhead\@empty
 \def\@oddfoot{\footnotesize\itshape
         {Submitted preprint} \hfill\today}%
 \let\@evenfoot\@oddfoot}
\makeatother

\usepackage{hyperref}
\usepackage{graphicx}
\usepackage{amssymb}

\usepackage{amsmath}
\usepackage{cleveref}

\usepackage{caption}

\usepackage{color}
\usepackage[table]{xcolor}
\usepackage{booktabs}
\usepackage{tabularx}

\newcolumntype{N}{>{\centering\arraybackslash}m{.28in}}

\journal{}

\biboptions{sort,comma,numbers}

\begin{document}

\begin{frontmatter}

\title{Deep convolutional autoencoder for cryptocurrency market analysis}

\author{Vladimir Puzyrev}
\address{Curtin University, Kent Street, Bentley, Perth, WA 6102, Australia \\ vladimir.puzyrev@ \{ \href{mailto:vladimir.puzyrev@gmail.com}{gmail.com}, \href{mailto:vladimir.puzyrev@curtin.edu.au}{curtin.edu.au} \} }

\begin{abstract}
This study attempts to analyze patterns in cryptocurrency markets using a special type of deep neural networks, namely a convolutional autoencoder. The method extracts the dominant features of market behavior and classifies the 40 studied cryptocurrencies into several classes for twelve 6-month periods starting from 15th May 2013. Transitions from one class to another with time are related to the maturement of cryptocurrencies. In speculative cryptocurrency markets, these findings have potential implications for investment and trading strategies.
\end{abstract}

\begin{keyword}
Machine learning \sep Autoencoder \sep Cryptocurrency \sep Bitcoin \sep Time varying \sep Market efficiency
\end{keyword}

\end{frontmatter}


\section{Introduction}

Cryptocurrencies have recently emerged as a digital alternative to traditional government-issued paper monies, secure electronic payment system, as well as financial and speculative assets. Cryptocurrencies are not controlled by central banks, behave differently to traditional currencies and other assets and possess several unique characteristics, which have attracted considerable attention in recent years not only from the media and investors, but also from the academic community. While being regarded by some categories of stakeholders as a technology that carries potential risks, cryptocurrencies have become commonly recognized as an attractive alternative investment for diversifying portfolio risks. In particular, Bitcoin, the most popular and largest by market capitalization cryptocurrency, has been shown to be able to serve as a hedge against stocks, currencies, gold, oil, and other financial assets \citep{dyhrberg2016hedging, guesmi2019portfolio, kang2019bitcoin}. Various altcoins have emerged in recent years, most of them created via initial coin offerings (ICOs). As of October 2019, there are more than 2300 cryptocurrencies, though Bitcoin represents 67\% and the next nine largest cryptocurrencies together represent 23\% of the total market capitalization (\href{http://coinmarketcap.com}{coinmarketcap.com} accessed on 27th October 2019).

Most academic studies to date have focused on Bitcoin, while research on other cryptocurrencies is quite limited with a few recent exceptions \citep{gkillas2018application, phillip2018new, wei2018liquidity}. Previous studies on the Bitcoin market showed that it is still in the early stages and its efficiency caused debates (\citealp{urquhart2016inefficiency}, \citealp{nadarajah2017inefficiency, bariviera2017inefficiency, tiwari2018informational}). While Bitcoin and several established cryptocurrencies have improved in terms of market efficiency in recent years, new cryptocurrencies have limited liquidity and exhibit strong signs of autocorrelation and non-independence \citep{wei2018liquidity, brauneis2018price}. \citet{corbet2018exploring} studied connections of cryptocurrencies to traditional assets and found evidence of strong interconnection of the cryptocurrencies with each other but their relative isolation from other assets and thus small sensitivity to market shocks. \citet{vidal2018semi} reported that Bitcoin has become more efficient over time in relation to its own events, while not being affected by monetary policy news. A recent study on interlinkages of cryptocurrencies by \citet{katsiampa2018volatility} also shows evidence of strong interdependencies in the cryptocurrency market and that the volatility and correlation between coins are responsive to major events. Most cryptocurrencies exhibit long memory, leverage, stochastic volatility, and heavy tailedness \citep{phillip2018new}. Evidence of short-term bubbles has been found in Bitcoin and Ethereum markets as well \citep{corbet2018datestamping}. \citet{baur2019bitcoin} studied seasonality patterns in Bitcoin prices and trading volume and found no persistent effects in returns across time, although there is a significant weekend effect in trading volume.

Prediction of Bitcoin and altcoin market movements is a challenging task. While empirical studies show that these markets are to some extent predictable with conventional technical analysis, high volatility and regular price jumps complicate the forecasting problem. The existence of "whales", who hold a significant quantity of existing Bitcoins and other crypto-assets, opens the door for price manipulation, especially on emerging altcoin markets. In the Bitcoin market, the volume was shown to be able to predict returns when the market is functioning around the normal mode, but not the volatility of returns \citep{balcilar2017can}. \citet{demir2018does} examined the prediction power of the economic policy uncertainty (EPU) index on the Bitcoin returns and concluded that Bitcoin has a hedging capability against the uncertainty. \citet{dastgir2019causal} established a bi-directional causal relationship between Bitcoin attention measured by the Google Trends search queries and returns. From the technical analysis perspective, the presence of strong nonlinearities in the returns and in the relationship between the returns and volume requires financial models that are able to capture nonlinearity and leverage effects. A possible solution to this problem is offered by deep learning.

Following the success in various fields of science and technology, deep neural networks (DNNs) have recently emerged as an efficient tool for analysis and forecasting financial data \citep{bao2017deep}. The main advantages of these methods are that they allow to detect and exploit the nonlinear dependencies in the data without specifying a particular model in advance and are able to discover both the low-level and high-level features by exploiting different layers of abstraction. This is of particular importance for financial large data sets where complex data interactions are currently difficult or impossible to specify in a full economic model \citep{heaton2016deep}. Moreover, deep learning methods are well-suited for handling large data sets, scale better with problem size compared to other machine learning techniques, and often can produce more useful results than standard financial tools.

The first applications of DNNs for cryptocurrency markets focus on Bitcoin price forecasting \citep{jang2018empirical, lahmiri2019cryptocurrency, han2019using} and employ recurrent neural networks, mainly based on the long short term memory (LSTM) model. Another type of DNNs called a convolutional neural network (CNN) is widely applied in classification problems but is much less common in financial time series forecasting, although recent studies \citep{borovykh2017conditional} show great promise of CNNs in time series analysis due to their ability to learn filters that represent repeating patterns and thus capture dependencies without the need for long historical data. CNNs can also be efficiently applied to clustering problems where their inherent property of highly nonlinear transformation allows to transform data with highly complex structure into more clustering-friendly representations \citep{min2018survey}.

In this study, I employ a novel data-driven feature extraction method using a special type of a deep CNN, namely a convolutional autoencoder, to analyze patterns in the behavior of the most important cryptocurrencies and shed some light on potential implications of this analysis for investment and trading strategies. The remainder of the paper is organized as follows. In Section 2, I describe the data used in this study and show some evidence of the correlation between the cryptocurrencies. The architecture of the neural network employed is described in Section 3. Section 4 shows the results of the analysis for the 40 largest cryptocurrencies. In the final section, some concluding remarks are given.

\section{Data and preliminary statistics}

The data used in this study is sourced from \href{http://coinmarketcap.com}{coinmarketcap.com} and includes the volume-weighted averages from different exchanges for the 40 largest cryptocurrencies measured by market capitalization. Altcoins that have market capitalization less than \$100,000,000 USD or are traded very little are excluded from this study. The full list of the cryptocurrencies used in this study and their current market shares are given in the Appendix. The data ranges from 15th May 2013 (new cryptocurrencies are added as they appear on market) to 15th May 2019 and includes three variables, namely, the daily closing price denominated in USD, the high-low price ratio that reflects both variance and the spread \citep{corwin2012simple}, and the trading volume. The logarithmic returns for currency $k$ are defined as
\begin{equation} \label{eq:Returns}
R_{k,t} = \mathbf{ln}(P_{k,t}) - \mathbf{ln}(P_{k,t-1}),
\end{equation}
where $P_{k,t}$ is the price value at time $t$.

\begin{figure}
\centering \makebox[\textwidth][c]{\includegraphics[width=1.0\linewidth]{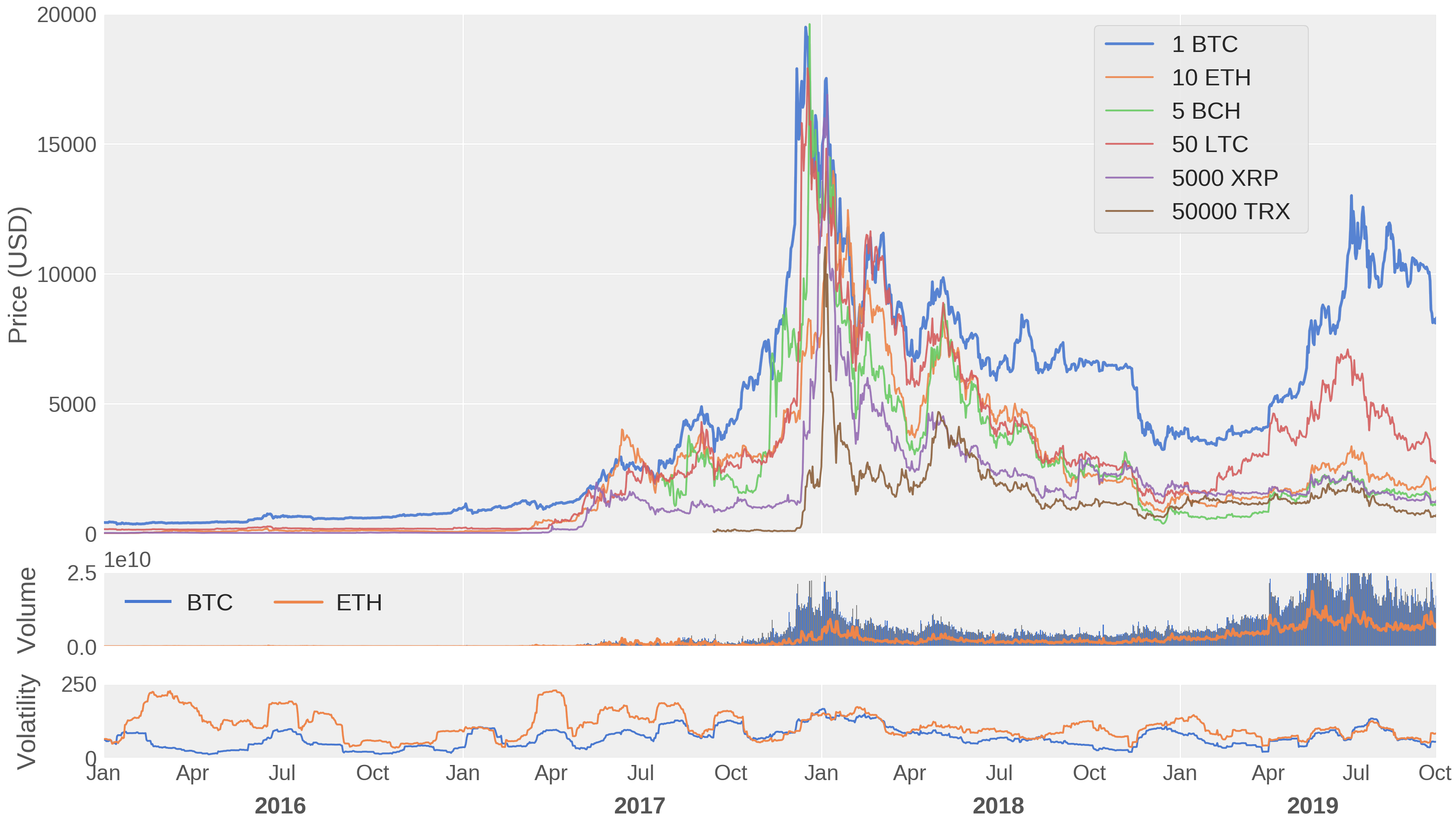}}
\caption{Daily closing prices for BTC, ETH, BCH, LTC, XRP, and TRX from 1st Jan 2016 to 1st October 2019.}
\label{Fig:CCplot1}
\end{figure}

Figure \ref{Fig:CCplot1} compares the daily closing prices for Bitcoin (BTC) and five major altcoins, namely Ethereum (ETH), Bitcoin Cash (BCH), Litecoin (LTC), Ripple (XRP), and TRON (TRX) from 1st Jan 2016 to 1st October 2019. All of them are characterized by sharp price changes with high transaction volumes. A simple visual inspection of the plots shows signs of a significant correlation between the cryptocurrency prices, as well as between BTC and ETH (the second largest cryptocurrency) trading volume and volatility. The volume reaches its peaks during the periods of extreme market movements, while the volatility remains very high for most of the largest cryptocurrencies, although slowly decreasing since early 2018.

\begin{figure}
\centering \includegraphics[width=1.0\linewidth]{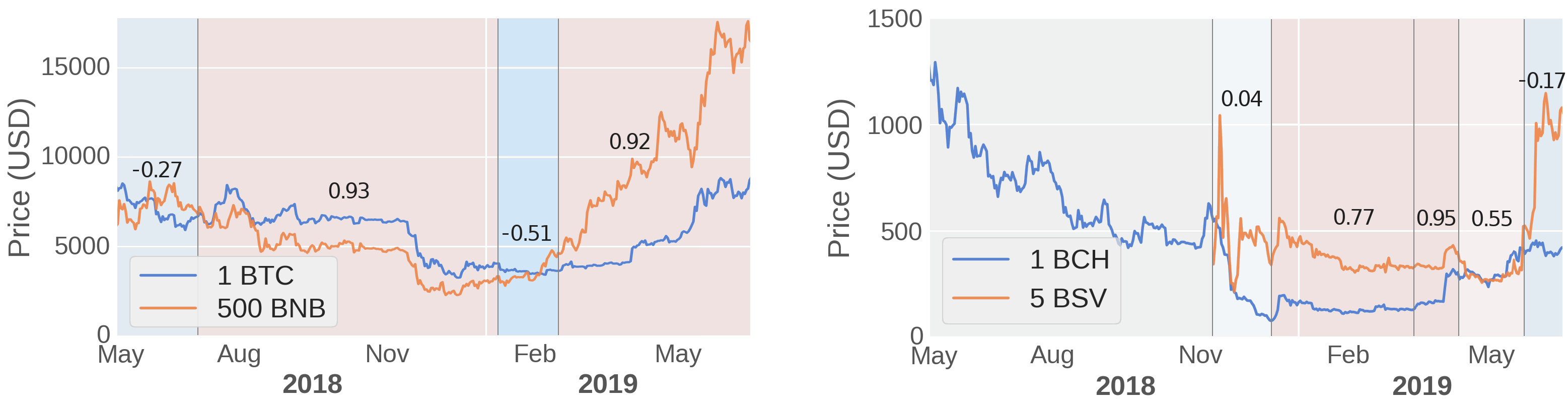}
\caption{Comparison of the daily closing prices of BTC/BNB and BCH/BSV since May 2018. The numbers denote the Pearson correlation coefficient, which measures the linear correlation between the prices for a given period. Negative and positive correlation coefficients are shown in light-blue and light-red colors, respectively.}
\label{Fig:CCcorre}
\end{figure}

At the same time, not all cryptocurrencies follow Bitcoin trends and those who exhibit strong correlation in one period may not correlate in another one. Two recent examples of this behavior are shown in Figure \ref{Fig:CCcorre}. The price of Binance Coin (BNB) experiences two periods of strong positive correlation and two periods of a weaker negative correlation with Bitcoin prices. Similar effects are observed for Bitcoin SV (BSV), a new cryptocurrency that stemmed from BCH in November 2018. As will be shown in Section 4, tests for the efficient market hypothesis for some cryptocurrencies exhibit similar fluctuations over time.

\section{Convolutional autoencoder}

A convolutional autoencoder (CAE, \citet{masci2011stacked}) is an unsupervised learning method, which is based on training the neural network to approximate the data by itself via a bottleneck structure. Unlike traditional supervised CNN-based approaches, it does not require large amounts of labeled training examples and can automatically learn discriminative features in data. Various autoencoders are widely applied nowadays for clustering \citep{guo2017deep, ghasedi2017deep, min2018survey} and anomaly detection \citep{chalapathy2019deep} tasks. 

CAE consists of two major parts, the \textit{encoder} $E_{\phi}$ that compresses the input $\mathbf{x}$ to lower-dimensional features $\mathbf{h}$ and the \textit{decoder} $D_{\theta}$ that takes the latent features as input and reconstructs the original data as closely as possible:
\begin{align} \label{eq:EncDecoder}
\begin{split}
& \mathbf{h} = E_{\phi}(\mathbf{x}), \\
& \tilde{\mathbf{x}} = D_{\theta}(\mathbf{h}).
\end{split}
\end{align}
Choosing the architecture of both the encoder and decoder as a deep CNN allows to learn hierarchical feature representations by exploiting deep features in time series data.

\begin{figure}
\centering \makebox[\textwidth][c]{\includegraphics[width=1.0\linewidth]{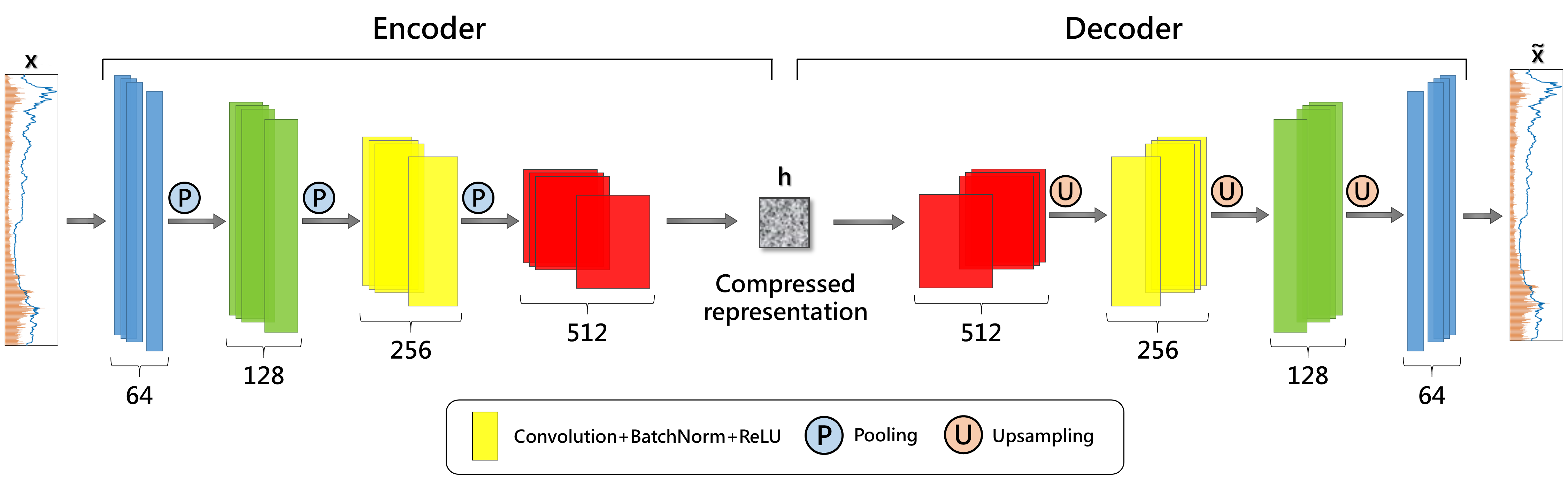}}
\caption{Architecture of the CAE used for cryptocurrency market data analysis. Color rectangles denote multichannel feature maps. The number of channels is shown at the bottom.}
\label{Fig:CAE}
\end{figure}

Figure \ref{Fig:CAE} shows the architecture of the CAE used in the following examples. It has 20.1 million trainable parameters and is composed of 25 layers: 12 convolutional/pooling layers in the encoder to build a deep representation of patterns in data, a feature layer of 10 units, and 12 convolutional/upscaling layers in the decoder. Each of the layers in the encoder and decoder networks contains between 64 and 512 filters for detecting various features in the input data.  Due to this hierarchical structure when the features extracted at the previous level become the input at the next level, the model is able to capture the short- and long-term market effects.

Batch normalization \citep{ioffe2015batch} is used after the convolutional layers for improving the training performance and regularization purposes. Rectified linear units (ReLU) are chosen as activation functions. Such 1D CNNs have shown good performance in sequential data processing where picking the most important features from segments of data regardless of the specific location of these features within the data is essential. The bottleneck structure of the CAE allows to capture the most important features of the input in the hidden feature layer $\mathbf{h}$. This layer has much smaller dimensionality than the original data, which leads to the creation of a compressed set of information in $\mathbf{h}$ from which the original data $\mathbf{x}$ is restored through linear and non-linear relationships. In other words, training of the CAE creates in $\mathbf{h}$ a more cost-effective representation of $\mathbf{x}$.

The training is performed in an unsupervised fashion using $\tilde{\mathbf{x}} = D_{\theta}( E_{\phi}(\mathbf{x}) ) = \mathbf{x}$. The parameters of the encoder and decoder networks, $\phi$ and $\theta$, respectively, are updated by minimizing the reconstruction error as follows:
\begin{equation} \label{eq:Optimization}
\min_{\phi, \theta} {L_r} = \min_{\phi, \theta} \frac{1}{n} \sum_{i=1}^{n} \left\lVert D_{\theta}({E_{\phi}({x_i})}) - x_i \right\rVert ^2_2.
\end{equation}
Here, the mean squared error (MSE) over the training dataset is used as the distance measure between the two variables. Nesterov-accelerated adaptive moment estimation (Nadam) algorithm \citep{dozat2016incorporating} is used for optimizing \eqref{eq:Optimization}.

Once the training is complete, the cryptocurrencies are split into several groups by applying the K-means clustering algorithm on the feature layer $\mathbf{h}$. For visualization, I employ the Principal Component Analysis (PCA) to extract the two dominant components in the feature vector. The implementation of these two algorithms is based on the scikit-learn Python package \citep{pedregosa2011scikit}. TensorFlow open-source library \citep{abadi2016tensorflow} is used for the CAE implementation.

\section{Results}

In this section, I employ the deep CAE to extract the dominant features of cryptocurrency market behavior and study interdependencies between the coins. Identification of such interdependencies is important for understanding the general market structure and its dynamic behavior. Previously, \citet{wei2018liquidity} categorized cryptocurrencies into 5 groups sorted by their Amihud illiquidity ratio, with most of the major coins belonging to the first group. This study shows how the cryptocurrencies can be categorized further using the features picked by a deep neural network from raw market data and how this categorization changes with time.

\begin{figure}
\centering \makebox[\textwidth][c]{\includegraphics[width=1.0\linewidth]{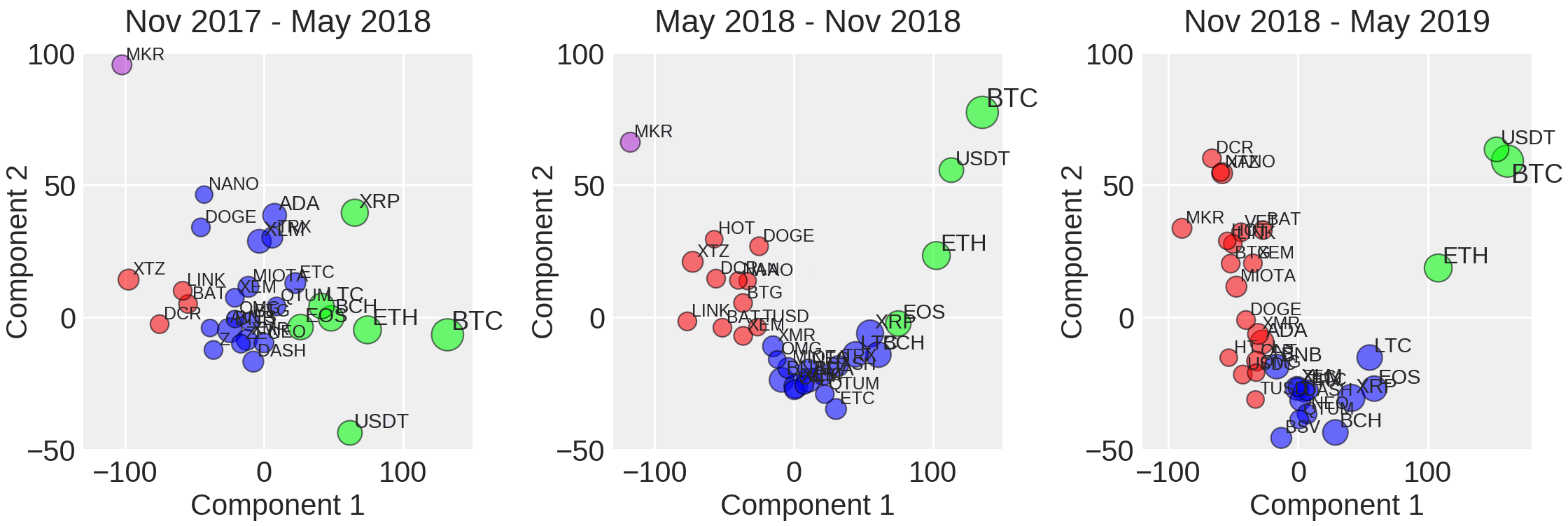}}
\caption{35 largest cryptocurrencies classified by the CAE for the periods of 16/11/2017 -- 15/05/2018, 16/05/2018 -- 15/11/2018, and 16/11/2018 -- 15/05/2019 using daily returns, high-low price ratio, and trading volume as input.}
\label{Fig:CAEresults1}
\end{figure}

Figure \ref{Fig:CAEresults1} shows the distribution of the dominant features of the cryptocurrencies over three consecutive 6-month periods. The input data includes returns, high-low price ratio, and trading volume for each cryptocurrency, thus allowing the neural network to learn by itself the relation between the volume and returns/volatility. For all studied periods, Bitcoin has characteristics different from most altcoins with Ethereum and Tether being the other two most extreme items. The next group consists of four major altcoins, namely EOS, Bitcoin Cash, Litecoin, and, to some extent, Ripple (the latter shows quite extreme characteristics in the first of the three shown periods). On the other side of the spectrum, we find Maker, a token based on the Ethereum blockchain aimed at stabilizing the value of Dai stablecoin. The distribution of the cryptocurrencies for the first period (November 2017 -- May 2018), which was characterized by extreme price changes, is quite different from the relatively stable second and third periods. This can be also attributed to the fact that the major coins have been moving towards becoming more efficient in recent years.

\begin{figure}
\centering \makebox[\textwidth][c]{\includegraphics[width=1.0\linewidth]{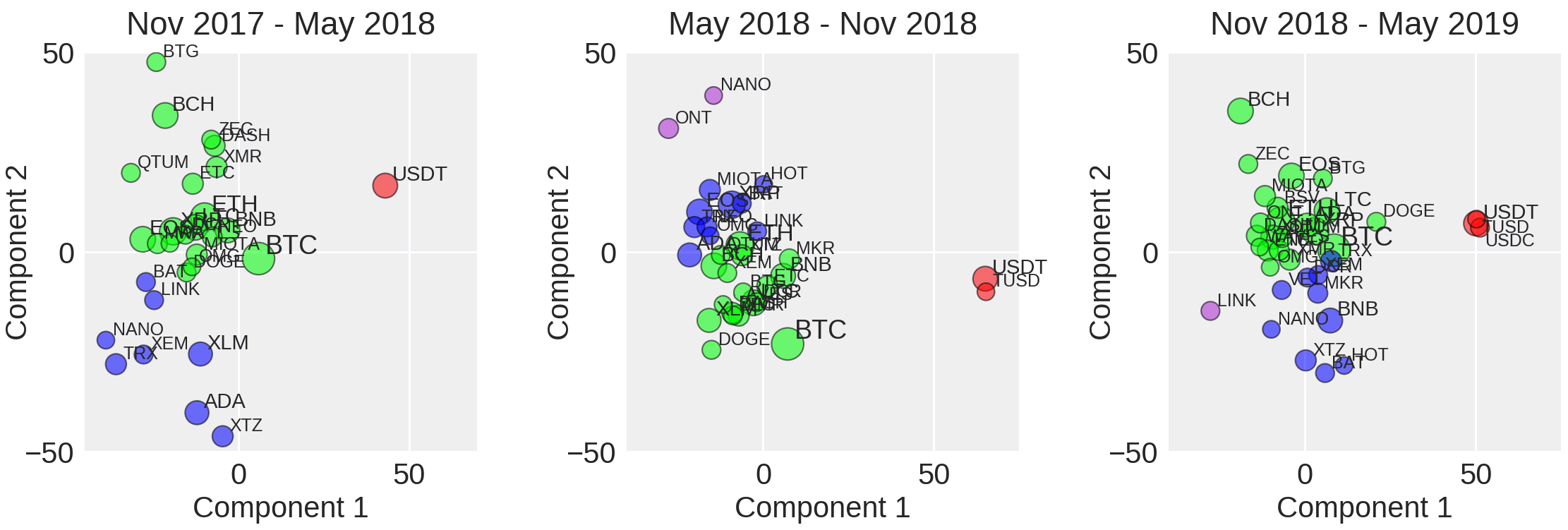}}
\caption{The same cryptocurrencies as in Figure \ref{Fig:CAEresults1} classified by the CAE using only daily returns as input.}
\label{Fig:CAEresults2}
\end{figure}

Figure \ref{Fig:CAEresults2} shows the CAE classification using only returns \eqref{eq:Returns} as input data. In this case, the first principal component directly reflects the volatility over a given period. Stablecoins meant to mirror the USD value (namely Tether in the first period and later TrueUSD and USD Coin) are clustered together separately from other coins. Another example worth mentioning is Chainlink (LINK), which does not exhibit extreme characteristics in the first two periods but becomes an outlier in the third one, which can be explained by its recent price rise. The opposite effect is observed for Dogecoin (DOGE), which experienced much fewer price fluctuations during the third period compared to the previous ones.

The observed changes in the cryptocurrency characteristics can be explained by their maturement and efficiency increase with time. Next, we consider how the results of the classical tests for the efficient market hypothesis exhibit dynamic fluctuations. Table \ref{Tab:Stats} shows the results for returns of ten major cryptocurrencies for three periods of 18 months each starting from 1st Jan 2015. The tests for randomness employed are the Ljung-Box (LB) test for no autocorrelation \citep{ljung1978measure}, the runs test \citep{wald1940test}, and the BDS test \citep{broock1996test}\footnote{The implementation relies on Python Statsmodels module by \citet{seabold2010statsmodels}. For the LB test, the lowest p-value of the first 10 lags is reported. The runs test uses the mean cutoff. In the BDS test, embedding dimensions vary from 2 to 5 and the mean p-value is reported; the distance parameter is set to 2.5.}. Data for the EUR/USD returns over the same periods is shown for comparison. As expected, as cryptocurrency markets mature, their efficiency characteristics experience significant changes. For the first five cryptocurrencies under consideration, the efficiency of returns increases with time suggesting that they are moving towards becoming more efficient. Bitcoin is known to be not directly affected by international monetary policy, although it actively responds to its own news, especially positive events, which can be explained by investors responding more quickly to positive news during the bull market periods in 2017 and 2019 \citep{vidal2018semi}. 

The changes in the cryptocurrency dominant features shown in Figure \ref{Fig:CAEresults2} are mainly related to the volatility, which is high for most of the established cryptocurrencies. For example, the average 30-day annualized volatility index since 2017 is 77.0\% for Bitcoin prices, while for the next six other altcoins (except Tether which is meant to mirror the USD value), this value is even higher and varies between 107.7\% and 148.1\%. For comparison, the EUR/USD average volatility index is 8.5\% over the same time period. New cryptocurrencies start from low market capitalizations and often exhibit high volatility as well (for example, Binance Coin and TRON average volatility during their first year was 182.6\% and 207.3\%, respectively; besides that, as the LB and BDS tests show in Table \ref{Tab:Stats}, their returns during this period do not follow a random walk).

\begin{table}[htbp]
\begin{center}
\caption{The Ljung-Box, runs, and BDS test for returns of ten cryptocurrencies (BTC, ETH, LTC, XRP, XLM, XMR, DASH, USDT, BNB, TRX) and Euro.}
\centering
\begin{tabular}{ c N N N N N N N N N }\toprule
 &\multicolumn{3}{c} {\textbf{1st Jan 2015 -}} & \multicolumn{3}{c} {\textbf{1st Jul 2016 -}} & \multicolumn{3}{c} {\textbf{1st Jan 2018 -}}\\
 &\multicolumn{3}{c} {\textbf{30th Jun 2016}} & \multicolumn{3}{c} {\textbf{31th Dec 2017}} & \multicolumn{3}{c} {\textbf{30th June 2019}}\\
 \cmidrule(lr){2-4}
 \cmidrule(lr){5-7}
 \cmidrule(lr){8-10}
 \textbf{Currency} & LB & Runs & BDS & LB & Runs & BDS & LB & Runs & BDS \\ 
 \cmidrule(lr){1-1}
 \cmidrule(lr){2-4}
 \cmidrule(lr){5-7}
 \cmidrule(lr){8-10}
 \textbf{BTC}  & 0.00 & 0.12 & 0.00 & 0.52 & 0.09 & 0.00 & 0.07 & 0.38 & 0.02 \\ 
 \textbf{ETH}  & 0.00 & 0.49 & 0.21 & 0.67 & 0.57 & 0.00 & 0.01 & 0.17 & 0.14 \\ 
 \textbf{LTC}  & 0.00 & 0.02 & 0.00 & 0.04 & 0.04 & 0.00 & 0.38 & 0.03 & 0.13 \\ 
 \textbf{XRP}  & 0.00 & 0.00 & 0.00 & 0.00 & 0.00 & 0.00 & 0.41 & 0.01 & 0.00 \\ 
 \textbf{XLM}  & 0.00 & 0.00 & 0.00 & 0.00 & 0.78 & 0.00 & 0.35 & 0.72 & 0.07 \\ 
 \textbf{XMR}  & 0.18 & 0.79 & 0.00 & 0.00 & 0.22 & 0.00 & 0.03 & 0.00 & 0.03 \\ 
 \textbf{DASH} & 0.35 & 0.03 & 0.00 & 0.14 & 0.31 & 0.00 & 0.02 & 0.04 & 0.03 \\ 
 \textbf{USDT} & 0.00 & 0.00 & 0.47 & 0.00 & 0.11 & 0.00 & 0.00 & 0.01 & 0.00 \\ 
 \textbf{BNB}  & --   & --   & --   & 0.00 & 0.71 & 0.00 & 0.01 & 0.58 & 0.00 \\ 
 \textbf{TRX}  & --   & --   & --   & 0.00 & 0.15 & 0.00 & 0.00 & 0.29 & 0.00 \\ 
 \textbf{EUR}  & 0.80 & 0.54 & 0.02 & 0.12 & 0.33 & 0.25 & 0.40 & 0.81 & 0.70 \\ 
 \bottomrule
\end{tabular}
\label{Tab:Stats}
\end{center}
\end{table}

\section{Conclusions}

In rapidly changing cryptocurrency markets, features learned by deep learning methods may potentially be more valuable and better suitable for analysis than hand-engineered features. In this study, a deep convolutional autoencoder is employed to automatically extract the dominant features of cryptocurrency behavior and conduct their classification for 6-month periods. The method successfully extracted the dominant features of the 40 studied cryptocurrencies and classified them into three or four classes, depending on the time period considered. The transition from one class to another with time is related to the maturement of cryptocurrency markets and changes in interdependencies between particular coins. 

Future research may examine whether similar patterns hold when technical indicators are added as inputs to the neural network. The proposed method can be extended to other financial markets and used as an additional tool for analyzing market structure, developing trading strategies and making investment decisions.


\bibliography{references}

\section*{Appendix}

Table \ref{Tab:CCList} lists the 40 cryptocurrencies chosen for this study, ranked by their market capitalization in USD. All together they currently represent around 96\% of the total market capitalization.

\begin{table}[htbp]
\begin{center}
\caption{40 cryptocurrencies used in this study. Market share is taken from \href{http://coinmarketcap.com}{coinmarketcap.com} on 27th October 2019. UNUS SED LEO, HedgeTrade, V Systems, Paxos Standard, and ABBC Coin were introduced recently and do not have enough historical data to be included in this study.}
\centering
\noindent
\begin{tabularx}{0.9\linewidth}{ >{\hsize=0.25\hsize}l >{\hsize=0.24\hsize}X >{\hsize=0.33\hsize}X | >{\hsize=0.25\hsize}l >{\hsize=0.24\hsize}X >{\hsize=0.33\hsize}X}\toprule
 \textbf{Name} & \textbf{Symbol} & \textbf{Market share (\%)} & \textbf{Name} & \textbf{Symbol} & \textbf{Market share (\%)} \\ 
Bitcoin & BTC & 67.45 &     Ethereum Classic & ETC & 0.22 \\
Ethereum & ETH & 8.08 &     Maker & MKR & 0.21 \\
XRP (Ripple) & XRP & 5.29 & USD Coin & USDC & 0.2 \\
Bitcoin Cash & BCH & 1.96 & Crypto.com Coin & CRO & 0.18 \\
Tether & USDT & 1.7 &       NEM & XEM & 0.16 \\ 
Litecoin & LTC & 1.59 &     Ontology & ONT & 0.16 \\
EOS & EOS & 1.31 &          Basic Attention Token & BAT & 0.14 \\
Binance Coin & BNB & 1.28 & Dogecoin & DOGE & 0.14 \\
Bitcoin SV & BSV & 1.08 &   Zcash & ZEC & 0.13 \\
Stellar & XLM & 0.55 &      VeChain & VET & 0.09 \\
TRON & TRX & 0.45 &         TrueUSD & TUSD & 0.08 \\
Cardano & ADA & 0.43 &      0x & ZRX & 0.07 \\
Monero & XMR & 0.41 &       Qtum & QTUM & 0.07 \\
Chainlink & LINK & 0.4 &    Decred & DCR & 0.06 \\
Huobi Token & HT & 0.35 &   Holo & HOT & 0.06 \\
IOTA & MIOTA & 0.31 &       Ravencoin & RVN & 0.06 \\
Dash & DASH & 0.27 &        Bitcoin Gold & BTG & 0.06 \\
NEO & NEO & 0.27 &          LUNA & LUNA & 0.05 \\
Cosmos & ATOM & 0.24 &      OmiseGO & OMG & 0.05 \\
Tezos & XTZ & 0.23 &        Nano & NANO & 0.05 \\
 \bottomrule
\end{tabularx}
\label{Tab:CCList}
\end{center}
\end{table}

\end{document}